\documentclass[letterpaper, 10 pt, conference]{ieeeconf} 
\IEEEoverridecommandlockouts                             

\usepackage{graphicx}
\usepackage{booktabs}
\usepackage{multirow}
\usepackage{colortbl}
\usepackage{xcolor}
\usepackage{amssymb}
\let\labelindent\relax 
\usepackage{enumitem}
\usepackage{threeparttable}
\usepackage[linesnumbered,ruled,vlined]{algorithm2e}

\usepackage[accsupp]{axessibility}

\definecolor{tabhl}{HTML}{E1F3F8}
\usepackage{hyperref}

\usepackage{cleveref}
\usepackage{orcidlink}

\usepackage{multibib}

\title{\LARGE \bf
APVI-SLAM: Real-Time Acoustic-Pressure-Visual-Inertial Localization and Photorealistic Mapping System in Complex Underwater Environment
}

\author{Hanwen Zhang$^{1}$, Yipeng Zhu$^{1}$, Xiaopeng Guo$^{1}$, Huajian Huang$^{2,\dag}$, Sai-Kit Yeung$^{3}$
\thanks{\textsuperscript{\dag} Corresponding Author}
\thanks{$^{1}$Hanwen Zhang, Yipeng Zhu and Xiaopeng Guo are with the Division of Integrative Systems and Design, Hong Kong University of Science and Technology (\scriptsize email: \{%
\href{mailto:hzhangfr@connect.ust.hk}{hzhangfr}, %
\href{mailto:yzhudg@connect.ust.hk}{yzhudg}, %
\href{mailto:xguoay@connect.ust.hk}{xguoay}%
\}@connect.ust.hk) 
}
\thanks{$^{2}$Huajian Huang is with the School of Interdisciplinary Science, Beijing Institute of Technology (\scriptsize email: huajian@bit.edu.cn)}
\thanks{$^{3}$Sai-Kit Yeung is with the Division of Integrative Systems and Design, the
Department of Computer Science and Engineering, and the Department of
Ocean Science, Hong Kong University of Science and Technology 
(\scriptsize email: saikit@ust.hk)}
}

\begin{document}

\maketitle
\thispagestyle{empty}
\pagestyle{empty}

%%%%%%%%%%%%%%%%%%%%%%%%%%%%%%%%%%%%%%%%%%%%%%%%%%%%%%%%%%%%%%%%%%%%%%%%%%%%%%%%
\begin{abstract}
Extreme subsea environments often cause severe feature de-gradation and estimator divergence in underwater visual-inertial SLAM. Although sensors like Doppler Velocity Logs (DVL) and pressure gauges provide auxiliary constraints, robust multi-sensor fusion during intermittent visual failure remains challenging. 
To address this, we present APVI-SLAM, a real-time multi-sensor fusion SLAM system that achieves both accurate underwater localization and photorealistic mapping. 
Our approach introduces a reliability-aware localization framework that dynamically reweights sensor estimators and employs a sliding-window freezing strategy to recover from tracking failures, substantially enhancing system robustness. Furthermore, for high-fidelity scenes reconstruction, we propose an efficient quadtree-guided mapping module that facilitates incremental water-medium modeling and 3D Gaussian optimization. Recognizing the lack of benchmark for underwater mapping evaluation, we also contribute a coral reef surveying dataset with synchronized multi-modality data. Extensive experiments on public and our proposed benchmarks demonstrate that APVI-SLAM achieves state-of-the-art localization and reconstruction quality at real-time speeds.
\end{abstract}

%%%%%%%%%%%%%%%%%%%%%%%%%%%%%%%%%%%%%%%%%%%%%%%%%%%%%%%%%%%%%%%%%%%%%%%%%%%%%%%%
\section{INTRODUCTION}

Underwater Visual–Inertial Simultaneous Localization and Mapping (SLAM) plays an important role in underwater robotics. Recovering photorealistic models of underwater environments facilitates marine exploration and survey tasks. 
However, rapid light attenuation, suspended particles, and ocean disturbances severely degrade imaging quality and limit the effective sensing range \cite{li2023review,wang2023overview,heshmat2025underwater}, often leading to unstable pose estimation and compromised reconstruction quality in Visual–Inertial (VI) SLAM systems. 

While integrating complementary measurements into factor graphs improves Visual-Inertial SLAM robustness \cite{xu2021underwater, rahman2022svin2, song2024turtlmap} in complex underwater environments, tightly-coupled multi-sensor estimators remain vulnerable. Since underwater visual feature loss is often persistent, visual degradation can corrupt the overall system estimation, even with covariance-based noise modeling \cite{xu2025aqua}. Moreover, complete visual dropouts force a brittle re-initialization of the VI component \cite{qin2018vins}, demanding time-consuming scale and covariance re-estimation. Beyond localization, existing underwater systems fail to reconstruct scenes with high photometric realism and structural consistency.

Recently, 3D Gaussian Splatting (3DGS) \cite{kerbl20233d} has emerged as an efficient and photorealistic scene representation. However, existing incremental Gaussian mapping methods\cite{matsuki2024gaussian,huang2024photo,wu2025vings} suffer from severe domain shift in underwater environments due to complex water-medium optical effects. SeaSplat \cite{yang2025seasplat} and WaterSplat \cite{li2025watersplatting} address this issue by modeling the water medium \cite{levy2023seathru}, while operating in an offline manner.
DUV-SLAM \cite{liu2025underwater} further improves incremental underwater 3D Gaussians mapping using deep neural network-based geometry estimation, yet its high computational cost still limits real-time deployment.

\begin{figure}[tb]
 \centering
  \includegraphics[width=0.95\linewidth]{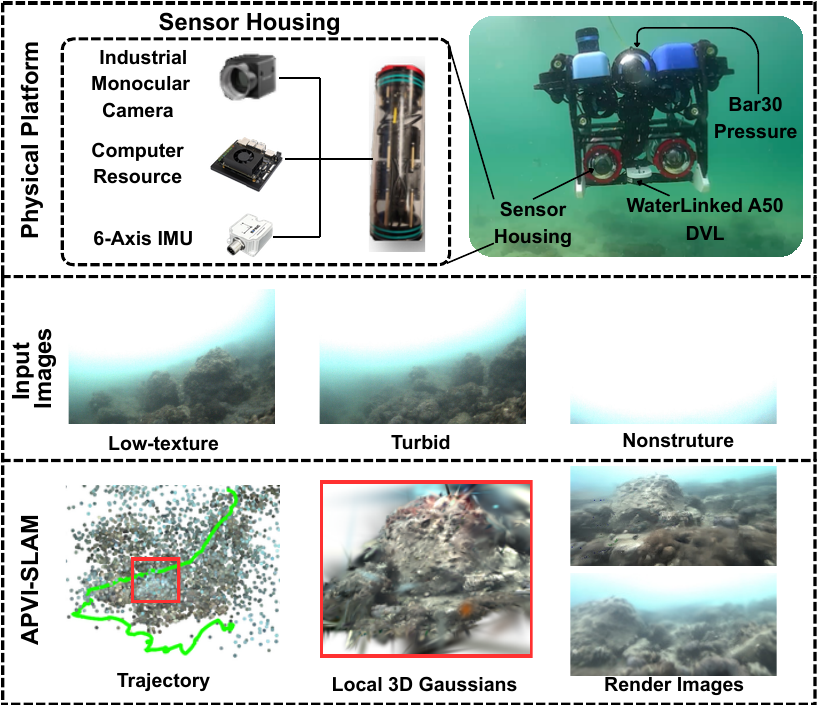}
  \vspace{-0.3cm}
  \caption{The proposed APVI‑SLAM enables real‑time multi‑sensor fusion for robust localization and photorealistic mapping in complex underwater environments, featuring deployment on an underwater robot and strong applicability to practical marine tasks.}
  \label{fig:teaser}
  \vspace{-0.7cm}
\end{figure}

In this paper, we present \textbf{APVI-SLAM}, a real-time multi-sensor fusion SLAM for robust localization and photorealistic mapping in complex underwater environments, as illustrated in Fig. \ref{fig:teaser}. Specifically, we introduce a reliability-aware sensor fusion localization framework that dynamically adjusts the weights of sensor estimators to prevent visual degradation from compromising the overall state estimation. This reliability awareness is driven by a coarse-to-fine self-discrimination mechanism designed to determine weight allocation under harsh conditions. To mitigate the high cost of re-initialization during total visual degradation, we propose a sliding-window freezing strategy that enables rapid recovery of visual tracking. Furthermore, to address complex underwater light transport and changing visibility, we develop a quadtree-guided 3D Gaussians mapping scheme featuring efficient water-medium modeling and scene optimization. Finally, we contribute a coral reef surveying dataset comprising synchronized multi-modality data with various degradation scenarios to bridge the gap of benchmark for underwater mapping evaluation. Experimental results demonstrate that APVI-SLAM achieves state-of-the-art performance in both localization accuracy and photorealistic mapping.

Overall, our work presents several key contributions, summarized as follows:
\begin{itemize}[leftmargin=2em]
\item We present APVI-SLAM, a novel multi-sensor fusion SLAM designed for real-time 
perception in complex underwater environment.
\item We propose a new reliability‑aware sensor fusion localization method that effectively enhances pose estimation robustness under estimator divergence.
\item We introduce a quadtree‑guided 3D Gaussians mapping scheme to efficiently refine spatial representation and recover photorealistic models even in turbid water conditions.
\item We construct the first coral reef surveying dataset with multi-modality data, benchmarking mapping performance in real-world underwater applications.
\item We implement the complete system in C++ and CUDA, achieving state-of-the-art performance on existing public and our newly proposed datasets.
\end{itemize}

\section{Related Works}
\subsection{Underwater Multi-Sensor Fusion SLAM}
Underwater Visual-Inertial SLAM faces significant challenges due to limited visibility and frequent feature degradation. To enhance localization robustness, many studies have adopted multi-sensor fusion frameworks that combine visual, inertial, acoustic, and pressure sensors. Early works relied on extended Kalman filtering for sensor fusion \cite{eustice2004visually, wang2025improved, kang2026hybrid}, while more recent methods employ factor-graph-based optimization \cite{ali2025advancing, guo2026robust, chen2026sonid}. Several Visual–Inertial systems \cite{rahman2022svin2, song2023acoustic} integrate acoustic data into classic VINS \cite{qin2018vins} system to improve accuracy. UVA \cite{xu2021underwater} tightly couples a DVL and stereo camera within a graph-based framework for accurate localization but suffers from frequent disturbances caused by intermittent visual degradation. AQUA-SLAM \cite{xu2025aqua} fuses inertial sensor further to utilize high-frequency information. These methods are sensitive to unstable visual inputs and don't support to reconstruct high‑fidelity maps. In contrast, our method introduces a novel reliability-aware sensor fusion localization method to enhance tracking module robustness. 
\subsection{Underwater Dense Mapping}
Previous underwater mapping frameworks \cite{song2024turtlmap,xu2025aqua,xu2021underwater} usually depend on expensive stereo camera setups and lack photorealistic mapping capability due to the pointcloud representation. Recently, 3D Gaussian Splatting \cite{kerbl20233d} emerged as an efficient representation for high‑fidelity scene reconstruction. While SeaSplat \cite{yang2025seasplat} and WaterSplatting\cite{li2025watersplatting} integrate 3DGS with water-medium models\cite{levy2023seathru} to achieve high-quality underwater reconstruction but lack incremental mapping capability. DUV-SLAM \cite{liu2025underwater} advances incremental mapping by employing an optical flow \cite{teed2021droid} front-end coupled with uncertainty-aware geometry. 
However, its inference speed limits real-time performance. Meanwhile, traditional feature‑based front‑ends \cite{huang2024photo} remain constrained by low-texture underwater environments, which yield sparse geometry prior and degrade optimization ability. 
To address these limitations, we propose a quadtree-guided 3D Gaussians mapping method to enable water medium modeling and incremental  optimization, enabling real-time photorealistic mapping in complex underwater environments.

\begin{figure*}[tb]
 \centering
 \vspace{-0.5cm}
  \includegraphics[width=0.85\linewidth]{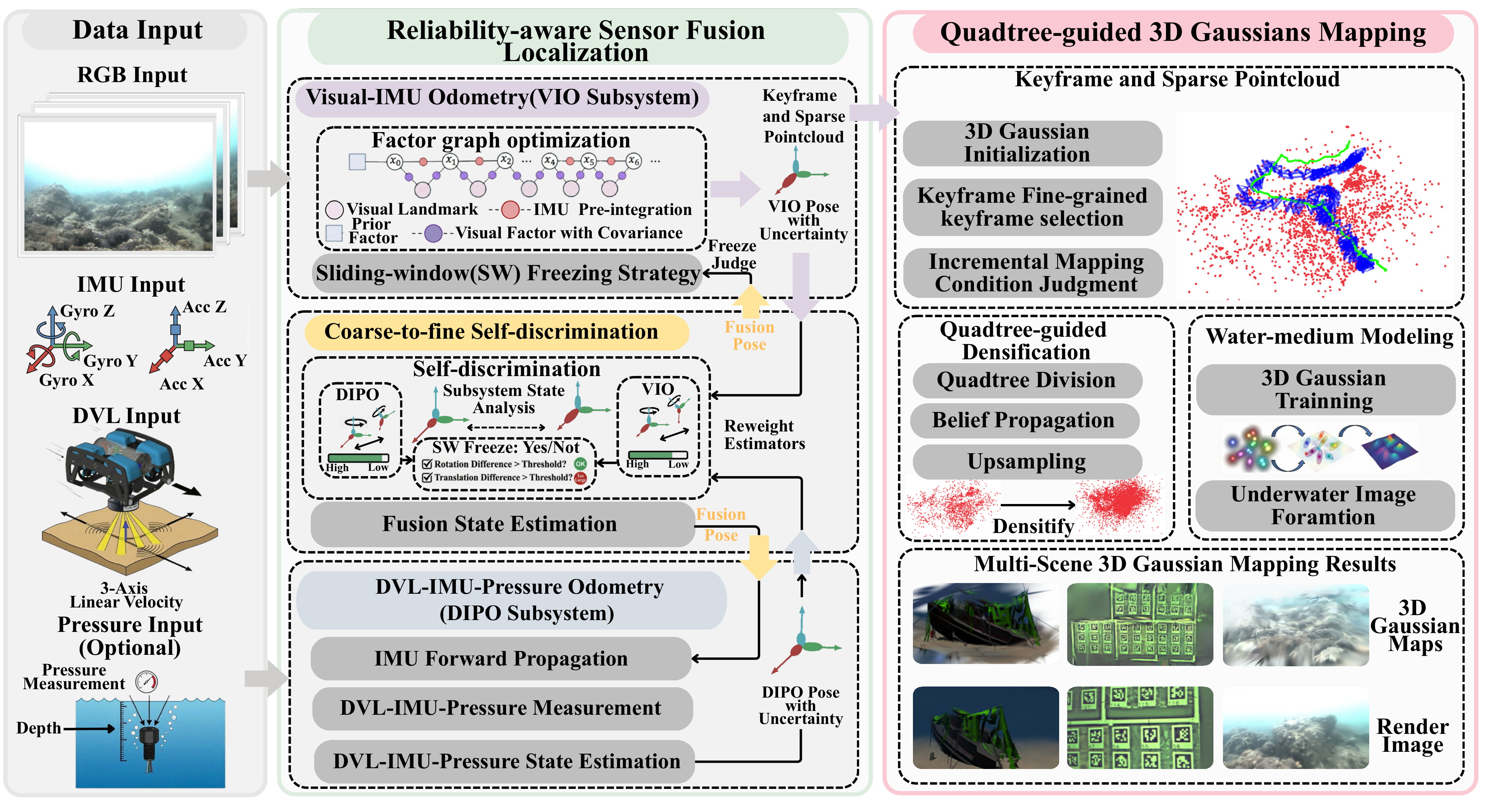}
  \caption{
  Overview of the proposed APVI-SLAM framework. Multi-sensor data, including RGB images, IMU, DVL, and pressure measurements, are integrated through a reliability-aware localization module to enhance tracking robustness in challenging underwater environments. These estimates subsequently drive a quadtree-guided 3D Gaussians mapping scheme, facilitating photorealistic underwater mapping.
  }
  \label{fig:pipeline}
   \vspace{-0.5cm}
\end{figure*}

\section{Methodology of APVI-SLAM}
The framework of APVI-SLAM is composed of two main components, including reliability-aware sensor fusion localization and quadtree-guided 3D Gaussians mapping, as illustrated in Fig. \ref{fig:pipeline}. The localization component takes RGB monocular images, IMU, DVL and pressure sensor measurements as input for pose estimations. 
We then reweight the DVL‑Inertial‑Pressure and Visual‑Inertial odometry estimations based on reliability awareness.
Given robust estimated poses, we recover high-quality underwater 3D Gaussians with efficient quadtree-guided densification.

\subsection{Preliminary}
\label{3-1}
The notations used in this paper are defined as follows: a scalar is represented by an italic lowercase letter $a$; a vector by a bold lowercase letter 
$\mathbf{a}$; a matrix by a bold uppercase letter $\mathbf{A}$; a coordinate frame by a typewriter font symbol $\mathtt{A}$.

\paragraph{State Estimation}
First, we define the world coordinate system $\mathtt{W}$ as the first frame IMU coordinate system aligned with the gravity system. The body coordinate $\mathtt{B}$ is the IMU frame. We use the prior extrinsic calibration to handle measurements taken in different sensor frames (DVL $\mathtt{D}$, pressure $\mathtt{P}$ and camera $\mathtt{C}$ coordinate). The state of the system at time i can be described as
\vspace{-0.1cm}
\begin{equation}
\footnotesize
\mathbf{x}_{i} \doteq [ \mathbf{R}_{i}, \mathbf{p}_{i}, \mathbf{v}_{i}, \mathbf{b}^g_{i}, \mathbf{b}^a_{i} ] \in \text{SO}(3) \times \mathbb{R}^{12}~.
\label{equa:eq-1}
\vspace{-0.1cm}
\end{equation}
where $\mathbf{R}_{i} \in \text{SO}(3)$ and $\mathbf{p}_{i}$ as the state rotation and position from body frame to world frame. i.e. $(\mathbf{R}_{\mathtt{W}\mathtt{B}_i}, \mathbf{p}_{\mathtt{W}\mathtt{B}_i}) = \mathbf{T}_{i} \in \text{SE}(3)$, while $\mathbf{v}_{i} \doteq {_{\mathtt{B}_i}\mathbf{v}} \in \mathbb{R}^3$ represents the linear velocity in body frame. $\mathbf{b}^g_{i}$ and $\mathbf{b}^a_{i}$ are the IMU gyroscope and accelerometer biases. 

\paragraph{Scene Representation based on 3D Gaussian Primitives}
3D Gaussian Splatting \cite{matsuki2024gaussian} represents a 3D scene using a set of spatially distributed anisotropic Gaussian primitives. Each of them are associated with point cloud position from tracking part as mean $\mathbf{o}\in\mathbb{R}^{12}$, scaling $\mathbf{s}\in\mathbb{R}^{3}$, density $\sigma\in\mathbb{R}^{1}$, spherical harmonic coefficients $\mathbf{SH}\in\mathbb{R}^{16}$ and covariances $\mathbf{\Sigma}\in\mathbb{R}^{3\times3}$. 

\subsection{Reliability-aware Sensor Fusion Localization}
\label{3-2}
To achieve robust pose estimation based on complementary DVL and pressure sensors, we propose a novel reliability-aware sensor fusion localization method designed to enhance estimation stability under intermittent visual degradation conditions. Specifically, we redistribute the potentially degraded visual observations and the complementary stable DVL, pressure sensor data, and adaptively reweight based two separate subsystems estimators.
For visual estimator, RGB images from camera and inertial data from IMU is input into the Visual-IMU subsystem named VIO and optimizes the system states using a factor‑graph‑based \cite{dellaert2017factor} method.
\paragraph{Visual-IMU Odometry (VIO) Subsystem}
\label{3-2-1}
Since underwater images often lack sufficient visual features, we employ traditional optical flow \cite{beauchemin1995computation} tracking as the front end of the VIO instead of descriptor‑based feature matching \cite{rublee2011orb}. 
Due to water background, exposure settings and camera noise, many feature tracks from front-end are unstable which can pose challenges for optimization. Therefore, instead of using a constant pixel noise \cite{qin2018vins}, we build a weighted information matrix $\Sigma_{\mathcal{V}}^{-1} = \left( \alpha \cdot \frac{n - n_{\min}}{n_{\max} - n_{\min}} + \beta \right) \mathbf{I}$ for feature points. $n$ is the number of times a feature point has been successfully tracked, $n_{\min}$ and $n_{\max}$ are the minimum and maximum tracking thresholds, respectively. 
The parameters $\alpha$ and $\beta$ are linear coefficients used to scale the influence of the tracking consistency, 
and $\mathbf{I}$ denotes the identity matrix. 
The back end of VIO performs nonlinear factor graph optimization over a sliding window to jointly estimate history, current states $^{VIO}\mathbf{x}_{i}$ and feature points $\left \{ \mathbf{o} \right \}$:
\vspace{-0.1cm}
\begin{equation}
\footnotesize
J(\mathbf{x}) = \sum \| \mathbf{r}_\mathcal{I}(\mathbf{x}) \|_{\Sigma_\mathcal{I}}^2 +
\sum \| \mathbf{r}_\mathcal{V}(\mathbf{x}) \|_{\Sigma_{\mathcal{V}}}^2
+ \cdots ~.
\vspace{-0.1cm}
\end{equation}
where $\mathbf{r}_\mathcal{I}$ constrains relative motion using preintegrated IMU measurements; $\mathbf{r}_\mathcal{V}$ penalizes the misalignment between 3D landmarks and their 2D projections; $\Sigma_\mathcal{I}^{-1}$ is the weighting information matrix derived from the propagation of IMU noise.
And then we output the current VIO frame state $^{VIO}\mathbf{x}_{i}$, the feature point tracking counts and stability, as well as the Hessian matrix of the 6-DOF state $^{VIO}\mathbf{H}_{i}$ from nonlinear optimization for reweighted visual-inertial estimator. To fuse complementary sensor measurement during visual unstable even degraded, we design the DIPO system to estimate the current state from DVL, IMU, and pressure sensors.

\paragraph{DVL-IMU-Pressure Odometry (DIPO) Subsystem}
The DIPO subsystem estimates the state at current time $\mathbf{x}_{i}$ from the last state $\mathbf{x}_{i-1}$ using an Error-State Kalman Filter \cite{sola2017quaternion} (ESKF). The filtering algorithm is motivated by the intuition that direct measurements of linear velocity and depth from DVL and pressure sensors. The motion equation based on IMU forward propagation needs to be constructed as follows:
\vspace{-0.1cm}
\begin{equation}
\footnotesize
        \mathbf{x}_{i} = \mathbf{x}_{i-1} \boxplus \left( \Delta t\, f(\mathbf{x}_{i-1}, \mathbf{u}_{i-1}, \mathbf{w}_{i-1}) \right)~,
\label{equa:eq-2}
\vspace{-0.1cm}
\end{equation}
where $\Delta t$ is IMU sampling period, $\boxplus$ define the generalized addition, and $f()$ based on IMU angular velocity and linear acceleration \cite{xu2021fast}.The adoption of the ESKF framework is motivated by the explicit spatial measurements obtained from the DVL and pressure sensors. Based on this observation model, the DVL measurement equation can be expressed as:
\vspace{-0.1cm}
\begin{equation}
\footnotesize
        \mathbf{z} = \mathbf{v}_{i} - \mathbf{v}_{i}^{m}, \mathbf{R}^{T}(\mathbf{p}_{i} - \mathbf{p}_{i-1}) - \Delta\mathbf{p}_{i-1,i}^{m}~.
        \label{equa:eq-3}
\vspace{-0.1cm}
\end{equation}
where $\mathbf{v}_{i}^{m}$ is the linear velocity changes observed by DVL are transferred to the body frame and $\Delta\mathbf{p}_{i-1,i}^{m}$ is the translation residual is obtained based on $\mathbf{v}_{i}^{m}$. The pressure observation residual is constructed by estimating the vertical (z-axis) displacement at the current time from the depth computed by the pressure sensor. 
Because of state-dependent and low-dimensional nature of the observations, ESKF is employed to estimate the system state at time $i$ based on the aforementioned motion and observation equations.
Meanwhile, the inverse of the covariance  of the pose (translation and roatation) $^{DIPO}\mathbf{H}_{i} = \mathbf{P}^{-1}_{DIPO} \in \mathbb{R}^{6\times6}$ obtained from the ESKF is extracted as a measure of estimation confidence and also reweights complementary DVL and pressure estimator.
Based on the redistributed sensor observations from the above process, we then perform dynamic reweighting of sensor estimators to achieve robust pose estimation. To achieve this, we first evaluate the stability of each subsystem based on coarse‑to‑fine self-discrimination mechanism to determine the assigned weights and fusion strategy.

\begin{algorithm}[t!]
\small
  \caption{Coarse-to-fine Self-discrimination}
  \label{alg:self_discrimination}
  \DontPrintSemicolon

  \KwIn{
    DIPO state ${}^{DIPO}\mathbf{x}_i$ and covariance ${}^{DIPO}\mathbf{H}_i$, \\
    VIO state ${}^{VIO}\mathbf{x}_i$ and Hessian ${}^{VIO}\mathbf{H}_i$, \\
    Number of feature points $n_{fp}$ and stable points $n_{sfp}$.
  }
  \KwOut{
    Final system state $\mathbf{x}_i$.
  }

  \tcp{Step 1: Coarse}
  \If{$n_{fp} > 50$ \textbf{and} $n_{sfp} > 0.2 n_{fp}$}{
    
    \tcp{Step 2: Medium}
    $\Delta \mathbf{x} \gets \|{}^{DIPO}\mathbf{x}_i - {}^{VIO}\mathbf{x}_i\|$\;
    $\Delta z \gets |{}^{DIPO}z_i - {}^{VIO}z_i|, z_i = x_i[6]$\; 
    
    \If{$\Delta \mathbf{x} < 30\text{cm}$ \textbf{and} $\Delta z < 30\text{cm}$ \textbf{and} $\Delta \theta < 10^\circ$}{
      
      \tcp{Step 3: Fine}
      Solve 6-DOF $\lambda$ from ${}^{VIO}\mathbf{H}_i$ based on Eq. \ref{equa:eq-6}\;
      \If{VIO is stable based on $\lambda$}{
        \Return Accepted VIO State ${}^{VIO}\mathbf{x}_i$\;
      }
    }
  }

  \tcp{DIPO Stability Check}
  \If{DIPO is Stable}{
    \Return State with Freeze\;
  }
  \Else{
    \Return State with Drop\;
  }

\end{algorithm}

\paragraph{Coarse-to-fine Self-discrimination Mechanism} The function of this module includes evaluating the reliability of the VIO and DIPO subsystem, determining whether the current frame needs to be frozen (as required by the sliding-window freezing strategy), and output the odometry after fusion based on dynamic weights. One part of the module’s input comes from the state $^{DIPO}\mathbf{x}_{i}$ and covariance $^{DIPO}\mathbf{H}_{i}$ of the DIPO subsystem. The other part comes from the state $^{VIO}\mathbf{x}_{i}$ and covariance $^{VIO}\mathbf{H}_{i}$ of the current frame in the VIO system. The time offset between the two subsystems can be aligned by IMU propagation. The module first evaluates the stability of the VIO system using a coarse-to-fine strategy shown in Alg. \ref{alg:self_discrimination}. When the coarse judgment is negative, we directly determine VIO fail without performing a fine-grained judgment. \textbf{Step 1}: it determines whether the number of front-end feature points $n_{fp}$ exceeds 50 and whether more than 20\% of them are stably tracked $n_{sfp} > 0.2n_{fp}$. A feature point is considered stable if it is successfully tracked across 10 consecutive frames. \textbf{Step 2}: it checks whether the euclidean distance offset between $^{DIPO}\mathbf{x}_{i}$ and $^{VIO}\mathbf{x}_{i}$ is less than 30cm, the deviation in the $z$-direction is less than 30cm and the Euler angle is less than 10\(^\circ\). \textbf{Step 3}: The stability of the VIO system is evaluated by analyzing the Hessian matrix $^{VIO}\mathbf{H}_{i}$ of the factor graph solved using the Levenberg-Marquardt method\cite{marquardt1963algorithm}. 6-DOF eigenvalues is solved: 
\begin{equation}
\footnotesize
        ^{VIO}\mathbf{H}_{i}=\begin{bmatrix}
  \mathbf{H}_{11} & \mathbf{H}_{12}  \\
  \mathbf{H}_{21} & \mathbf{H}_{22}
\end{bmatrix}, \mathbf{\lambda_{1}}=eigen(\mathbf{H_{11}}), \mathbf{\lambda_{2}}=eigen(\mathbf{H_{22}})~,
        \label{equa:eq-6}
\end{equation}
According to \cite{benaych2016lectures}, the eigenvalues distribution of a randomly constructed real symmetric matrix with finite dimensions can be approximated with a semicircle distribution. By modeling the Hessian distribution on a simulation dataset, the thresholds $\lambda^{thre}_{1}$, $\lambda^{thre}_{2}$ are analytically determined as the lower edges of the distribution's support. Any eigenvalue falling below these boundaries indicates parsing degradation and VIO failure.
After self-discrimination, we perform reweighting of the VIO and DIPO estimators to obtain the fused state based on ESKF:
\begin{equation}
\footnotesize
\mathbf{x}_i =
\begin{cases}
^{DIPO}\mathbf{x}_i, & \text{if VIO fails}, \\
^{DIPO}\mathbf{x}_i + \mathbf{K}\left(^{VIO}\mathbf{x}_i - ^{DIPO}\mathbf{x}_i\right), & \text{otherwise}.
\end{cases}
\label{equa:eq-7}
\end{equation}
where $\delta \mathbf{x_i}$ is the error update variable, $\mathbf{K}$ is the Kalman gain. 
The inverse of $^{VIO}\mathbf{H}_{i}$ and $^{DIPO}\mathbf{H}_{i}$ are used as noise parameters for reweighting. As the DIPO system does not suffer from degeneracy, we assess its reliability based only on outlier rejection and the properties of its Hessian matrix $^{DIPO}\mathbf{H}_{i}$. When the VIO subsystem fails, the stable fusion output is marked with a "freeze" flag, and the non-stable output is marked with a "drop" flag.

\begin{figure}[t]
	\centering
	\includegraphics[width=0.95\linewidth]{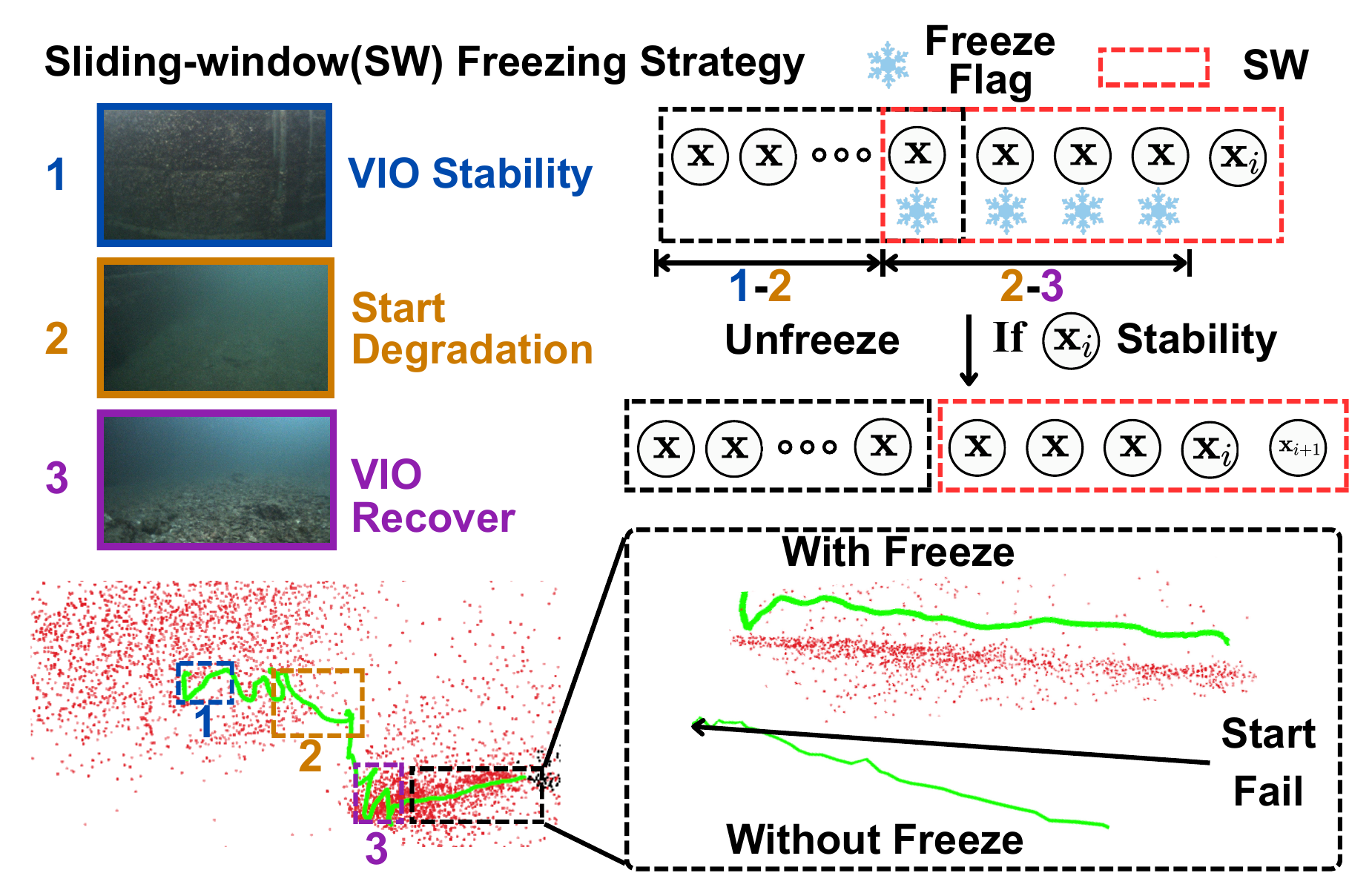}
    \vspace{-0.4cm}
	\caption{Sliding-window freezing strategy for efficient re-initialization.}
	\label{fig:sw_freeze_module}
     \vspace{-0.5cm}
\end{figure}

\paragraph{Sliding-window Freezing Strategy}
The purpose of this strategy is to efficient re-initialization of the VIO subsystem by utilizing the prior information from the stable DIPO subsystem.
Stable fusion output $\mathbf{x}_i$ with a freeze flag is maintained within the sliding window and used from factor graph optimization, whereas unstable results $\mathbf{x}_i$ with a drop flag are utilized solely as odometry outputs and excluded from the factor graph optimization, as illustrated in Fig. \ref{fig:pipeline}. This means that the factor graph preserves trustworthy state variables when VIO degrades, allowing the VIO subsystem to rapidly resume state solving after visual features reappear. Specifically, when the VIO subsystem has failed shown in Fig. \ref{fig:sw_freeze_module} 1-2, $\mathbf{x}_i$ with a freeze flag aim to keep this prior such as scale and state in the optimization process. During factor graph optimization, the keyframes marked with a "freeze" flag, as well as the feature points first observed by these keyframes, are kept fixed. As the visual degradation gradually diminishes shown in Fig. \ref{fig:sw_freeze_module} 2-3, the factor graph quickly converges without the need for repeated initialization. Once the VIO subsystem recovers normal operation, the "freeze" state of the historical keyframes is released.

\subsection{Quadtree-guided 3D Gaussians Mapping}
\label{3-3}
Intuitively, to avoid generating invalid photometric maps from degraded visual information during VIO failures, the mapping module selectively integrates localization data based on the system state and feature reliability. Only when the localization module is in a stable tracking state, the current keyframe $\mathbf{T}_{\mathtt{C}\mathtt{W}_{i}}$ pose and sparse pointclouds with high confidence explained in Sec. \ref{3-2-1} $\left \{ \mathbf{o} \right \}_{s}$  are fed into the Gaussian mapping system.
\paragraph{3D Gaussians Rendering based on Water-Medium Modeling}
Each Gaussian primitive with mean $\mathbf{o}$ and covariance $\mathbf{\Sigma}$ is transformed to current keyframe as $\mathbf{o}^{\mathtt{C}} = \mathbf{R}_{\mathtt{C}\mathtt{W}} \mathbf{o} + \mathbf{p}_{\mathtt{C}\mathtt{W}}$, $\mathbf{\Sigma}^{\mathtt{C}} = \mathbf{R}_{\mathtt{C}\mathtt{W}} \mathbf{\Sigma} \mathbf{R}_{\mathtt{C}\mathtt{W}}^{\top}$. 
Projection to 2D using intrinsics $\mathbf{K}$ yields $\boldsymbol{\mu} = \pi(\mathbf{o}^{\mathtt{C}})$ and $\mathbf{\Sigma}' = \mathbf{J} \mathbf{\Sigma}^{\mathtt{C}} \mathbf{J}^{\top}$ with Jacobian $\mathbf{J}$. Pixel radiance $\mathbf{C}(\mathbf{u})$ is computed via $\alpha$-blending. 
The final observed color $\mathbf{I}(\mathbf{u})$ applies the water-medium model\cite{akkaynak2018revised} with attenuation $\beta^{D}$, backscatter $\beta^{B}$, and background color $\mathbf{B}_{\infty}$:
\begin{equation}
\footnotesize
\mathbf{I}^{'}(\mathbf{u}) = \mathbf{C}(\mathbf{u}) \cdot e^{-\beta^{D} z(\mathbf{u})} + \mathbf{B}_{\infty} \cdot \left(1 - e^{-\beta^{B} z(\mathbf{u})}\right),
\label{eq:water_model}
\end{equation}
where depth $z(\mathbf{u})$ is obtained from the corresponding Gaussian.
Water-medium parameters are jointly optimized online, and the learning rate is doubled when the VIO subsystem is detected to have failed, in order to accelerate convergence. Moreover, the sparsity of feature points often results in weak-texture or turbid regions. To bridge this gap, we introduce a lightweight quadtree-guided densification scheme.

\paragraph{Lightweight Quadtree-guided Densification}
Pixels that share the same quadtree block in the intensity image also belong to the same block in the depth map\cite{wang2018quadtree}. Therefore, by dividing the image into a three-level quadtree($16\times16$, $8\times8$, $6\times6$) based on intensity, pixels within each level exhibit similar texture and depth characteristics. We then sample depth based on photometric cost $V(\mathbf{u}, d)$ for pixel $\mathbf{u}$:
\begin{equation}
\footnotesize
V(\mathbf{u}, d) = \frac{1}{N - 1} 
\sum_{i = 0}^{N - 2} 
\left| \mathbf{I}_{N-1}(\mathbf{u}) - \mathbf{I}_k \!\left( \pi(\mathbf{u}, d, \mathbf{T}_{k,N-1}) \right) \right|,
\end{equation}
where $\mathbf{I}(\cdot)$ is the intensity of a specific pixel in image $\mathbf{I}$, a sequence of 10 keyframes from the sliding windows in the VIO subsystem $(\mathbf{I}_0, \ldots, \mathbf{I}_{N-1})$. $\pi(\cdot)$ warps the coordinate $\mathbf{u}$ from image $\mathbf{I}_{N-1}$ to $\mathbf{I}_{i}$ given the depth $d$. we choose 128 depth hypothesis (i.e., $\left \{ d_{0},\dots,d_{63}   \right \}$)  that are equally spaced from 0.10–25\,m.  Based on photometric cost $V(\mathbf{u}, d)$, the depth of quadtree-selected pixels is estimated based on belief propagation $\mathbf{B}_{u_k}(d)$ from its 4-neighbor $\mathcal{N}(u_k)$:
\begin{equation}
\footnotesize
\mathbf{B}_{u_k}(d) = \mathbf{I}(u_k, d) + \sum_{s \in \mathcal{N}(u_k)} \bigl(m_{s \rightarrow u_k}(d)\bigr),
\end{equation}
where $m_{s \rightarrow u_k}$ is the message passing from pixel $\mathbf{s}$ to $\mathbf{u}_{k}$, the depth $d^*$ that minimizes the belief vector is selected as the estimation of the pixel. To fill in the missing regions and obtain a consistent depth, we perform depth upsampling\cite{engel2013semi}. During the Gaussian primitive densification process, we first track the sparse point cloud $\left \{ \mathbf{o} \right \}_{s}$ and locate their corresponding quadtree levels in the current frame. For these positions, additional densification is unnecessary. For other regions, we project the top-left representative pixel $\mathbf{u}_q$ of each quadtree block into the world coordinate system $\mathbf{T }_{\mathtt{W}\mathtt{C}_{i}} \cdot \pi^{- 1}(\mathbf{u}_q, d^*)$ to generate the densified point set $\left \{ \mathbf{o} \right \}_{d}$. Both $\left \{ \mathbf{o} \right \}_{d}$ and 
$\left \{ \mathbf{o} \right \}_{s}$ constitute all 3D Gaussians.

\begin{table}[t] 
    \centering
    \caption{Quantitative comparison of ATE (m) on the Tank dataset.}
    \vspace{-0.35cm}
    \label{tab:tank}
    \renewcommand\arraystretch{1.1}
    \setlength{\tabcolsep}{2pt} 
    \begin{threeparttable}
    \resizebox{0.95\linewidth}{!}{
    \begin{tabular}{l|c|ccc|ccc|cc|c}
        \toprule
        \multirow{2}{*}{Method} & \multirow{2}{*}{Sensors} & \multicolumn{3}{c|}{Structure} & \multicolumn{3}{c|}{HalfTank} & \multicolumn{2}{c|}{Whole} & \multirow{2}{*}{Avg.} \\
        \cmidrule(lr){3-5} \cmidrule(lr){6-8} \cmidrule(lr){9-10}
        & & Easy & Medium & Hard & Easy & Medium & Hard & Medium & Hard & \\
        \midrule
        ORB3 & M & 0.12 & - & - & 3.74 & - & - & - & - & -\\
        ORB3 & M+I & - & - & 3.25 & - & - & - & - & - & -\\
        ORB3 & S+I & 0.28 & 3.30 & 2.93 & 2.21 & 0.70 & 1.15 & 0.68 & 2.37 & 1.70 \\
        \midrule
        VINS-F. & M+I & 0.56 & - & - & 94.5 & - & - & - & - & - \\
        VINS-F. & S+I & 0.22 & - & - & 29.8 & - & - & 13.1 & - & - \\
        \midrule
        UVA & S+D & 0.18 & 0.49 & 0.43 & 1.12 & 0.26 & 0.37 & 0.27 & 0.30 & 0.43\\
        AQUA & S+I+D & \cellcolor{tabhl}\textbf{0.07} & 0.18 & 0.50 & \cellcolor{tabhl}\textbf{0.28} & 0.29 & 0.36 & 0.52 & \cellcolor{tabhl}\textbf{0.22} & 0.30\\
        \midrule
        \textbf{Ours} & M+I+D+P & 0.11 & \cellcolor{tabhl}\textbf{0.17} & \cellcolor{tabhl}\textbf{0.24} & \cellcolor{tabhl}\textbf{0.28} & \cellcolor{tabhl}\textbf{0.17} & \cellcolor{tabhl}\textbf{0.33} & \cellcolor{tabhl}\textbf{0.26} & 0.23 & \cellcolor{tabhl}\textbf{0.22} \\
        \bottomrule
    \end{tabular}
    }
    \begin{tablenotes}
            \tiny
            \item \colorbox{tabhl}{Bold} indicate the best performance. M: Mono, S: Stereo, I: IMU, D: DVL, P: Pressure. 
            \item ORB3 and VINS-F. stand for ORB-SLAM3 and VINS-Fusion.
        \end{tablenotes}
         \end{threeparttable}
    \vspace{-0.3cm}
\end{table}

\section{Experiments}
\subsection{Experiment Setup}
\paragraph{Datasets}
While the Tank dataset \cite{xu2025tank} provides multi-sensor data with marker-based ground truth under adverse conditions, it lacks the complexity required for mapping evaluation. To bridge this gap, we developed a high-fidelity simulation benchmark and a real-world coral reef surveying dataset. The simulation dataset, built via the UUV Simulator \cite{manhaes2016uuv}, includes 6 sequences across Ship, Seabed, and Dam environments with each scenario featuring both clear and degraded visibility. For field validation, we curated a coral reef surveying dataset using a customized BlueROV equipped with a forward industrial camera, 6‑axis IMU, Bar50 pressure sensor and WaterlinkA50 DVL. This dataset captures low-texture, turbid, and non-structural underwater environments, comprising 8 sequences with an average length of 2k frames.

\paragraph{Metrics} 
We use PSNR, SSIM, and LPIPS \cite{zhang2018unreasonable} to evaluate the photorealistic mapping quality. We also report computing resources by showing the tracking FPS, rendering FPS, and GPU memory usage. To evaluate the camera pose, we use the average absolute and relative trajectory error (ATE/RPE RMSE). 
Due to the significant noise levels in the raw data, our evaluation on the real-world dataset is restricted to qualitative comparisons, as quantitative metrics may not reliably reflect reconstruction quality.

\begin{figure}[t!]
 \centering
  \includegraphics[width=0.95\linewidth]{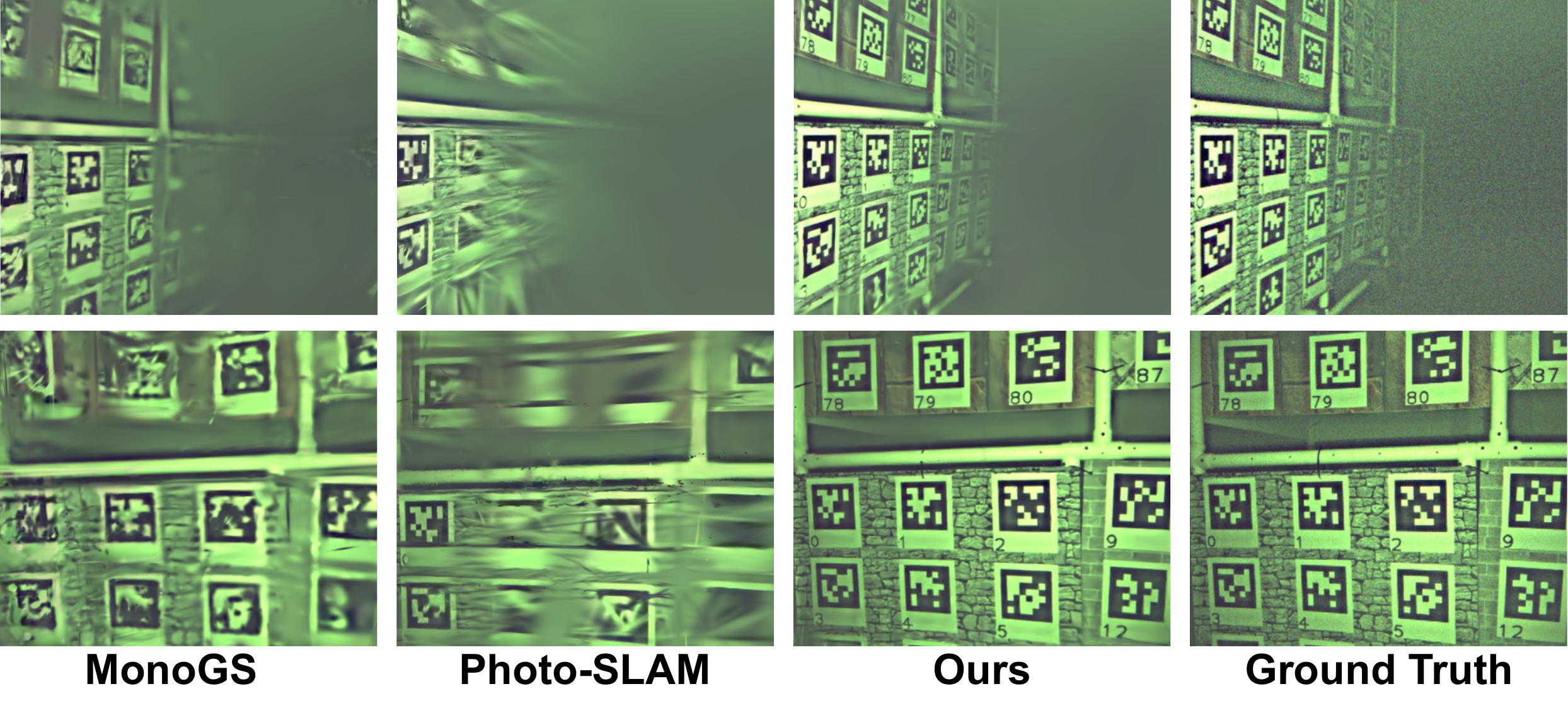}
  \vspace{-0.50cm}
  \caption{Qualitative results on the Tank dataset. Our method achieves high-fidelity scene reconstruction with sharp details, while others exhibit significant blurring.}
  \label{fig:tank_render}
  \vspace{-0.65cm}
\end{figure}

\begin{table*}[t!]
  \centering
  \caption{Quantitative comparison results on both easy and hard sequences of simulation dataset.  Our method shows \colorbox{tabhl}{SOTA} performance. 
  “$\times$” denotes no view rendering support.
  }
  \vspace{-0.35cm}
  \label{tab:simulation_easy}
  \renewcommand\arraystretch{1.2}
  \resizebox{0.85\textwidth}{!}{
  \begin{tabular}{l|cccc|cc|ccc|cccc}
      \toprule
      \multirow{2}{*}{Easy Seq.} & \multicolumn{4}{c|}{Sensors} & \multicolumn{2}{c|}{Localization} & \multicolumn{3}{c|}{Mapping} & \multicolumn{4}{c}{Resources} \\
      \cmidrule(lr){2-5} \cmidrule(lr){6-7} \cmidrule(lr){8-10} \cmidrule(lr){11-14}
      & Mono & IMU & DVL & Press. & ATE $\downarrow$ & RPE $\downarrow$ & PSNR $\uparrow$ & SSIM $\uparrow$ & LPIPS $\downarrow$ & Op. Time $\downarrow$ & T-FPS $\uparrow$ & R-FPS $\uparrow$ & GPU $\downarrow$ \\
      \midrule
      ORB-SLAM3  & \checkmark & \checkmark & & & 11.820 & 7.893 & $\times$ & $\times$ & $\times$ & 3.043 & 51.447 & $\times$ & $\times$ \\
      VINS-Fusion& \checkmark & \checkmark & & & 25.208 & 4.657 & $\times$ & $\times$ & $\times$ & 2.303 & \cellcolor{tabhl}\textbf{100.00} & $\times$ & $\times$ \\
      Go-SLAM    & \checkmark & \checkmark & & & 26.378 & 53.453 & 18.999 & 0.681 & 0.537 & \cellcolor{tabhl}\textbf{2.297} & 65.430 & 10.160 & \cellcolor{tabhl}\textbf{4 GB} \\
      Photo-SLAM & \checkmark & & & & 7.146 & 5.370 & 28.990 & 0.803 & 0.324 & 3.170 & 35.979 & 582.463 & 5 GB \\
      MonoGS     & \checkmark & & & & 486.311 & 249.677 & 27.010 & 0.788 & 0.349 & 27.554 & 2.880 & 2.254 & \cellcolor{tabhl}\textbf{4 GB} \\
      VINGS  & \checkmark & \checkmark & & & 498.284 & 398.275 & 17.656 & 0.558 & 0.540 & 10.940 & 26.333 & 332.165 & 5 GB \\
      \textbf{Ours} & \checkmark & \checkmark & \checkmark & \checkmark & \cellcolor{tabhl}\textbf{5.699} & \cellcolor{tabhl}\textbf{3.810} & \cellcolor{tabhl}\textbf{31.233} & \cellcolor{tabhl}\textbf{0.863} & \cellcolor{tabhl}\textbf{0.260} & 2.303 & \cellcolor{tabhl}\textbf{100.00} & \cellcolor{tabhl}\textbf{874.09} & \cellcolor{tabhl}\textbf{4 GB} \\
      \bottomrule
      \toprule
      \multirow{2}{*}{Hard Seq.} & \multicolumn{4}{c|}{Sensors} & \multicolumn{2}{c|}{Localization} & \multicolumn{3}{c|}{Mapping} & \multicolumn{4}{c}{Resources} \\
      \cmidrule(lr){2-5} \cmidrule(lr){6-7} \cmidrule(lr){8-10} \cmidrule(lr){11-14}
      & Mono & IMU & DVL & Press. & ATE $\downarrow$ & RPE $\downarrow$ & PSNR $\uparrow$ & SSIM $\uparrow$ & LPIPS $\downarrow$ & Op. Time $\downarrow$ & T-FPS $\uparrow$ & R-FPS $\uparrow$ & GPU $\downarrow$ \\
      \midrule
      ORB-SLAM3 & \checkmark & \checkmark  & & & 36.093 & 30.877 & $\times$ & $\times$ & $\times$ & 5.363 & 52.610 & $\times$ & $\times$ \\
      VINS-FUSION & \checkmark & \checkmark & & & 185.820 & 68.977 & $\times$ & $\times$ & $\times$ & \cellcolor{tabhl}\textbf{4.213} & \cellcolor{tabhl}\textbf{100.000} & $\times$ & $\times$ \\
      Go-SLAM & \checkmark & \checkmark & & & 413.222 & 154.992 & 21.633 & 0.583 & 0.615 & 5.053 & 67.510 & 9.887 & \cellcolor{tabhl}\textbf{4 GB} \\
      Photo-SLAM & \checkmark & & & & 404.759 & 14.477 & 30.483 & 0.817 & 0.429 & 5.513 & 35.880 & 934.799 & 5 GB \\
      MonoGS & \checkmark & & & & 501.531 & 218.740 & 27.350 & 0.717 & 0.444 & 22.017 & 4.180 & 3.447 & 5 GB \\
      VINGS & \checkmark & \checkmark & & & 633.000 & 356.792 & 21.443 & 0.738 & 0.507 & 12.713 & 30.551 & 370.715 & 7 GB \\
      \textbf{Ours} & \checkmark & \checkmark & \checkmark & \checkmark & \cellcolor{tabhl}\textbf{5.937} & \cellcolor{tabhl}\textbf{3.857} & \cellcolor{tabhl}\textbf{32.390} & \cellcolor{tabhl}\textbf{0.843} & \cellcolor{tabhl}\textbf{0.328} & \cellcolor{tabhl}\textbf{4.213} & \cellcolor{tabhl}\textbf{100.000} & \cellcolor{tabhl}\textbf{1035.497} & 6 GB \\
      \bottomrule
  \end{tabular}    
}
\vspace{-0.25cm}
\end{table*}

\begin{figure*}[tb]
 \centering
  \includegraphics[width=0.95\textwidth]{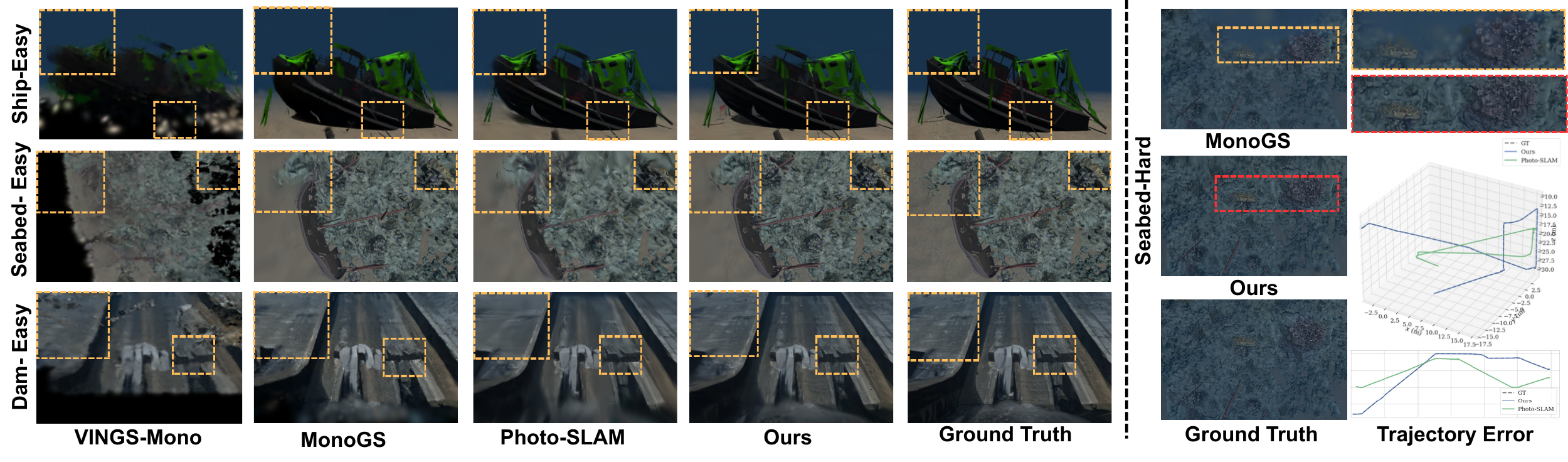}
  \caption{Left: Qualitative results on the easy sequences of simulation dataset. Our method demonstrates superior texture clarity and finer geometric details compared to existing baselines. Right: Qualitative results and trajectory visualization on the Seabed-Hard of simulation dataset. Photo-SLAM failed due to visual degradation. Our method provides richer texture details under low‑visibility conditions.
 }
  \label{fig:simulation_easy}
  \vspace{-0.40cm}
\end{figure*}

\paragraph{Baselines} 
We evaluate our method against state-of-the-art classical SLAM systems, including VINS-Fusion \cite{qin2018vins} and ORB-SLAM3 \cite{campos2021orb}. Comparisons are also conducted with underwater multi-sensor fusion frameworks such as UVA \cite{xu2021underwater} and AQUA-SLAM \cite{xu2025aqua}, which utilize factor-graph optimization but lack high-fidelity mapping capabilities. 
Since UVA is not open-sourced and AQUA-SLAM lacks a monocular configuration, we compare them on the Tank dataset using the original data from their original papers. Furthermore, we consider recent neural and Gaussian-based SLAM methods, including Go-SLAM \cite{zhang2023go}, Photo-SLAM \cite{huang2024photo}, and MonoGS \cite{matsuki2024gaussian}. To ensure a comprehensive evaluation, we include VINGS \cite{wu2025vings}, which incorporates IMU factors for photorealistic reconstruction.

\paragraph{Implementation} 
Our system is implemented in C++ and CUDA, leveraging LibTorch for both optimization modules and neural components to facilitate efficient GPU acceleration and flexible tensor computation. All experiments are conducted on a workstation equipped with an Intel i7-14700K CPU and an NVIDIA RTX 4070 GPU. Except for UVA and AQUA-SLAM, whose results are taken from their original publications, all other baselines are evaluated using their official implementations.

\subsection{Evaluation of Tank Public Dataset}
On the Tank dataset, our method achieves SOTA localization performance, as shown in \cref{tab:tank}. 
Compared to the most relevant underwater SLAM, AQUA-SLAM, 
our approach reducing position error by 26.7\% on average, while
significantly outperformng it as environmental complexity increases and visual degradation intensifies, notably in Structure Hard (0.24m vs. 0.50m) and WholeTank Medium (0.26m vs. 0.52m). Qualitatively, Fig. \ref{fig:tank_render} shows our method reconstructs high-fidelity scenes, whereas MonoGS and Photo-SLAM suffer from severe blurring and artifacts. These results validate the effectiveness of our method in mitigating underwater visual degradation while maintaining high-precision tracking and mapping.

\subsection{Evaluation of Our New Simulation Dataset}
As demonstrated in \cref{tab:simulation_easy}, our method consistently achieves superior performance across localization accuracy, mapping quality, and computational efficiency. In easy sequences with high visibility, MonoGS and VINGS still exhibit higher drift. 
In hard sequences, all baseline methods except ours fail to maintain accurate localization due to severe visibility reduction and feature degradation. 

In term of mapping, baselines struggle to produce complete reconstructions, leaving missing regions with no valid renderings. In contrast, our method  delivers photorealistic, detailed textures even under extreme degradation, as illustrated in Fig. \ref{fig:simulation_easy}. This resilience is primarily attributed to the robustness of the localization module and an efficient recovery mechanism that facilitates rapid re-initialization following periods of environmental degradation.

\subsection{Mapping Evaluation of Real Coral Surveying Dataset}
\begin{figure}
 \centering
  \includegraphics[width=\linewidth]{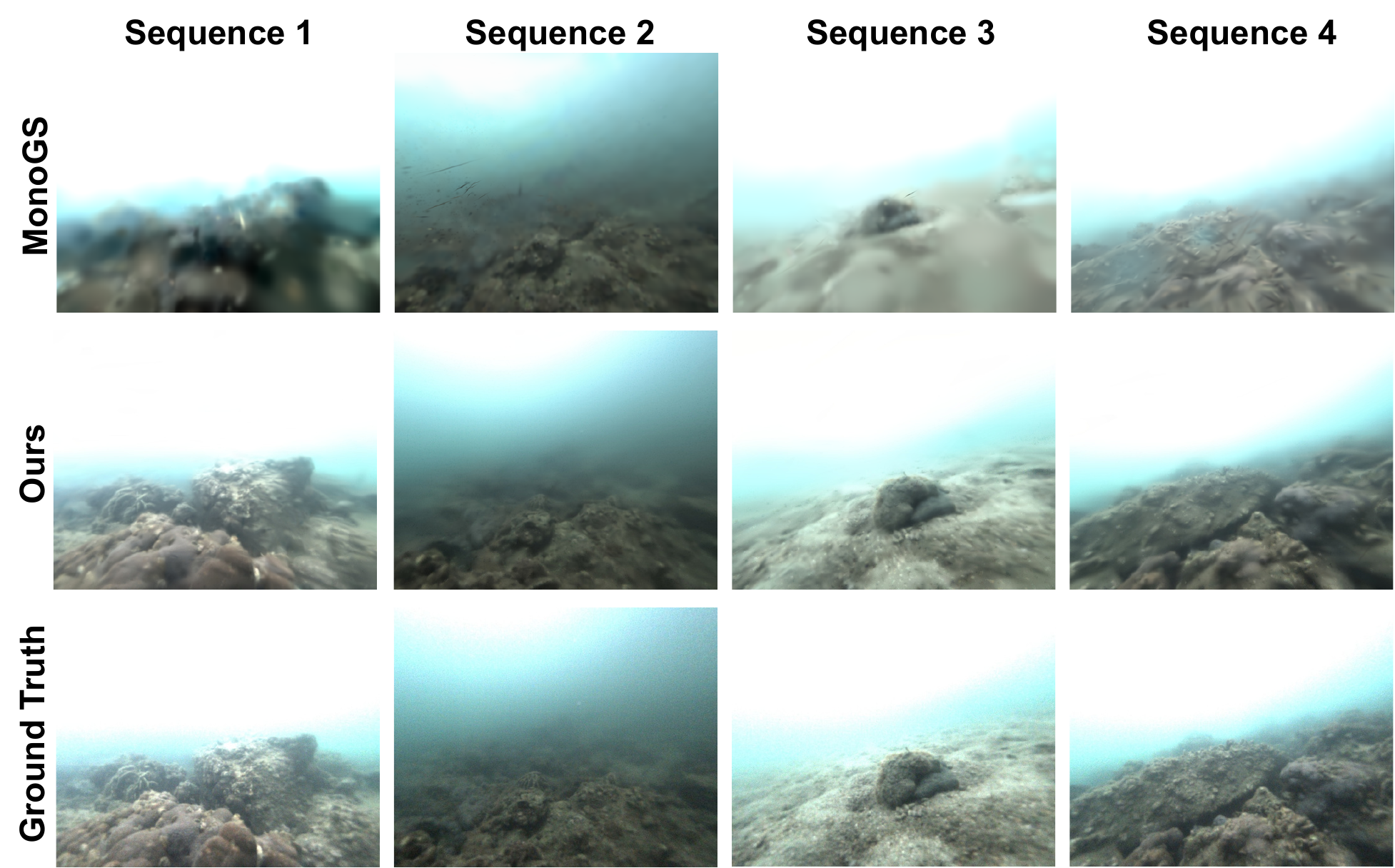}
  \vspace{-0.70cm}
  \caption{Qualitative results on the coral reef surveying dataset. Our method achieves high-fidelity mapping quality across different sequences.}
  \label{fig:real_reander}
    \vspace{-0.30cm}
\end{figure}

In field scenarios, constrained illumination and limited underwater visibility introduce substantial sensor noise and severe visual degradation. 
Such adverse conditions hinder Photo-SLAM’s ability to track ORB features, resulting in tracking failure and incomplete reconstructions. VINGS fails to initialize stable trajectories under these circumstances. 
Despite MonoGS manages to produce a complete map, its limited localization accuracy leads to blurred texture rendering (24.14 PSNR). 
Conversely, our method exhibits remarkable robustness to real-world challenges, delivering photorealistic reconstructions (29.05 PSNR) with fine-grained and spatially consistent details, shown in Fig. \ref{fig:real_reander}.

\subsection{Ablation Experiment}
Table \ref{tab:tank_ab} evaluates the incremental contribution of each sensor modality.  While the visual-inertial (VI) estimator fails under degraded conditions (Medium and Hard), the DVL-inertial configuration maintains robust meter-level accuracy, underscoring the necessity of acoustic-inertial sensing in turbid water. 
The integration of pressure measurements further refines the DVL-inertial estimate (0.31m vs. 0.38m) by providing critical vertical constraints. 
Ultimately, our full configuration consistently yields the superior localization (0.22m) compared to any sensor subset. 
These results demonstrate that the reliability-aware fusion effectively leverages cross-modal complementarities, achieving robust performance across varying environmental complexities.

\begin{table}[t]
    \centering
    \caption{Ablation study on sensor settings for the Tank dataset.}
    \vspace{-0.15cm}
    \label{tab:tank_ab}
    \renewcommand\arraystretch{1.1}
    \setlength{\tabcolsep}{2.5pt} 
    \resizebox{1.0\linewidth}{!}{ 
    \begin{tabular}{l|c|ccc|ccc|cc|c}
        \toprule
        \multirow{2}{*}{Settings} & \multirow{2}{*}{Sensors} & \multicolumn{3}{c|}{Structure} & \multicolumn{3}{c|}{HalfTank} & \multicolumn{2}{c|}{Whole} & \multirow{2}{*}{Avg.} \\
        \cmidrule(lr){3-5} \cmidrule(lr){6-8} \cmidrule(lr){9-10}
        & & Easy & Medium & Hard & Easy & Medium & Hard & Medium & Hard & \\
        \midrule
        1 & M+I & 0.12 & - & - & 3.76 & - & - & - & - & - \\
        2 & I+D & 0.24 & 0.28 & 0.28 & 0.84 & 0.36 & 0.35 & 0.38 & 0.32 & 0.38 \\
        3 & I+D+P & 0.15 & 0.23 & 0.26 & 0.62 & 0.27 & 0.36 & 0.32 & 0.25 & 0.31 \\
        \midrule
        \textbf{Default} & M+I+D+P & \cellcolor{tabhl}\textbf{0.11} & \cellcolor{tabhl}\textbf{0.17} & \cellcolor{tabhl}\textbf{0.24} & \cellcolor{tabhl}\textbf{0.28} & \cellcolor{tabhl}\textbf{0.17} & \cellcolor{tabhl}\textbf{0.33} & \cellcolor{tabhl}\textbf{0.26} & \cellcolor{tabhl}\textbf{0.23} & \cellcolor{tabhl}\textbf{0.22} \\
        \bottomrule
    \end{tabular}
    }
    \begin{tablenotes}
        \tiny
        \item[*] M: Mono, I: IMU, D: DVL, P: Pressure. 
        % \item[*] \textbf{Def.} denotes our default full-sensor configuration.
    \end{tablenotes}
    \vspace{-0.4cm}
\end{table}

\begin{table}[t]
    \centering
    \caption{Ablation study about the sliding-window freezing strategy on the Tank dataset.}
    \vspace{-0.30cm}
    \label{tab:supp_abu_sw}
    \renewcommand\arraystretch{1.2}
    \resizebox{0.95\linewidth}{!}{
    \begin{tabular}{c|c|cc|ccccc|c}
\toprule
\multirow{2}{*}{Metrics}                                                 & \multirow{2}{*}{Method}                                         & \multicolumn{2}{c|}{Structure} & \multicolumn{3}{c}{HalfTank}                 & \multicolumn{2}{c|}{WholeTank} & \multirow{2}{*}{Avg.} \\ \cline{3-9}
                                                                         &                                                                 & Medium         & Hard          & Easy   & Medium & \multicolumn{1}{c|}{Hard}  & Medium         & Hard          &                       \\ \hline
\multirow{2}{*}{\begin{tabular}[c]{@{}c@{}}Success \\ Rate\end{tabular}} & \begin{tabular}[c]{@{}c@{}}Ours w/o \\ sw freezing\end{tabular} & 60\%           & 0\%           & 100\%  & 10\%   & \multicolumn{1}{c|}{0\%}   & 60\%           & 20\%          & 21\%                  \\ \cline{2-10} 
& \textbf{Ours}  & \cellcolor{tabhl}\textbf{100\%}   & \cellcolor{tabhl}\textbf{100\%}        & \cellcolor{tabhl}\textbf{100\%}  & \cellcolor{tabhl}\textbf{100\%}  & \multicolumn{1}{c|}{\cellcolor{tabhl}\textbf{100\%}} & \cellcolor{tabhl}\textbf{100\%}         & \cellcolor{tabhl}\textbf{100\%}        & \cellcolor{tabhl}\textbf{100\%}                \\ \hline
\multirow{3}{*}{\begin{tabular}[c]{@{}c@{}}Re-init.\\ Time\end{tabular}} & Reference                                                       & 37.24s         & 76s           & \cellcolor{tabhl}\textbf{32.23s} & 7.13s  & \multicolumn{1}{c|}{127s}  & \cellcolor{tabhl}\textbf{11.14s}         & \cellcolor{tabhl}\textbf{8.03s}            & 37.24s                \\ \cline{2-10} 
& \begin{tabular}[c]{@{}c@{}}Ours w/o \\ sw freezing\end{tabular} & 46.37s          & x             & 35.95s  & 9.29s  & \multicolumn{1}{c|}{x}     & 12.21s         & x             & x                     \\ \cline{2-10} 
& \textbf{Ours}    & \cellcolor{tabhl}\textbf{10.24s}          & \cellcolor{tabhl}\textbf{10.98s}         & 32.52s  & \cellcolor{tabhl}\textbf{6.95s}   & \multicolumn{1}{c|}{\cellcolor{tabhl}\textbf{50.31s}} & 11.25s          & 8.43s          & \cellcolor{tabhl}\textbf{10.24s}                 \\ \bottomrule
\end{tabular}
}
 \begin{tablenotes}
        \tiny
        \item[*] “$\times$” denotes the system fails to reinitialized.  
    \end{tablenotes}
    \vspace{-0.4cm}
\end{table}

\begin{table}[t] 
    \centering
    \caption{Ablation study on quadtree-guided densification and water-medium modeling(W-M) on simulation dataset.}
    \vspace{-0.3cm}
    \label{tab:mapping_ab}
    \renewcommand\arraystretch{1.1}
    \setlength{\tabcolsep}{2.8pt} 
    \resizebox{0.95\linewidth}{!}{
    \begin{tabular}{lcc|ccc|ccc}
        \toprule
        \multicolumn{3}{c|}{Settings} & \multicolumn{3}{c|}{Easy Sequence} & \multicolumn{3}{c}{Hard Sequence} \\ 
        \cmidrule(lr){1-3} \cmidrule(lr){4-6} \cmidrule(lr){7-9}
        \# & Densification \tnote{1} & W-M\tnote{2} & PSNR$\uparrow$ & SSIM$\uparrow$ & R-FPS$\uparrow$ & PSNR$\uparrow$ & SSIM$\uparrow$ & R-FPS$\uparrow$ \\ 
        \midrule
        1 & Grad. & $\times$ & 29.93 & 0.852 & \cellcolor{tabhl}\textbf{971.2} & 29.35 & 0.819 & \cellcolor{tabhl}\textbf{1100.4} \\
        2 & Random+Grad. & $\times$ & 30.50 & 0.857 & 921.6 & 30.28 & 0.825 & 1067.2 \\
        3 & Quad.+Grad. & $\times$ & 30.96 & 0.859 & 958.7 & 31.94 & 0.829 & 1095.3 \\
        4 & Grad. & $\checkmark$ & 30.18 & 0.854 & 943.3 & 31.75 & 0.824 & 1079.3 \\ 
        \midrule
        5 SeaSplat & Grad. & $\checkmark$ & 29.95  & 0.858  & 959.2  & 31.67  &0.822  &1059.8 \\ 
        \midrule
        \textbf{Default} &Quad.+Grad.  & $\checkmark$ & \cellcolor{tabhl}\textbf{31.23} & \cellcolor{tabhl}\textbf{0.863} & 874.1 & \cellcolor{tabhl}\textbf{32.39} & \cellcolor{tabhl}\textbf{0.843} & 1035.5 \\ 
        \bottomrule
    \end{tabular}
    }
    \begin{tablenotes}
        \tiny
        \item[*] Grad.: The original 3DGS gradient-based densification method.
        Quad.: Quadtree-guided densification method.
    \end{tablenotes}
    \vspace{-0.4cm}
\end{table}

We evaluates the sliding-window freezing strategy through Success Rate and Re-initialization (Re-init.) Time in Tab. \ref{tab:supp_abu_sw}. Our method achieves a 100\% success rate across all sequences; in contrast, omitting this strategy leads to initialization failures in "Hard" environments due to the lack of stable visual anchors. 
Regarding efficiency, our approach minimizes Re-init. Time by maintaining a valid state during visual gaps, outperforming the baseline that allows the VIO state to drift. These results confirm that the sliding-window freezing strategy effectively bridges periods of severe degradation, enabling rapid VIO recovery while the DIPO subsystem ensures continuous, high-frequency pose outputs.

We also evaluate our lightweight quadtree-guided densification and water-medium modeling schemes in Tab. \ref{tab:mapping_ab}. 
Due to the sparsity of feature points, a simple random densification step also becomes necessary, as illustrated in Tab. \ref{tab:mapping_ab} (1)and(2). From Tab. \ref{tab:mapping_ab} (2)to(3), the lightweight quadtree‑guided densification strategy reconstructs the scene more efficiently, reducing the rendering time from 971.24 to 958.74. By observing Tab. \ref{tab:mapping_ab} (3)and(4), the impact of water-medium modeling is particularly pronounced in challenging sequences, where it mitigates distance-dependent color shifts to yield a significant PSNR gain from 0.25 to 2.37. 
In Tab. \ref{tab:mapping_ab} (5), our method outperforms the offline SeaSplat\cite{yang2025seasplat} with state-of-the-art water modeling, which uses the same poses and iterations as our method to ensure fairness, due to our efficient mapping strategy. These results confirm that the proposed scheme significantly elevates reconstruction fidelity while maintaining a near-constant computational overhead.

\section{CONCLUSIONS}
In this paper, we propose a novel real-time multi-sensor fusion SLAM system called APVI-SLAM for robust underwater localization and photorealistic mapping. 
We propose a reliability-aware sensor fusion localization framework and sliding-window freezing strategy to enhance tracking module robustness and mitigate high cost re-initialization process. Furthermore, we introduce a quadtree-guided 3D Gaussians mapping scheme to reconstruct photorealistic underwater scenes. Finally, we collect a coral reef monitoring dataset to address the lack of mapping focus in existing datasets. Extensive experiments have demonstrated that APVI-SLAM significantly outperforms existing SLAMs while maintaining real-time speed in complex underwater environments. Future work will address remaining challenges, including dynamics object, water‑medium removal, and the enhancement of 3D Gaussian geometric structures.
%%%%%%%%%%%%%%%%%%%%%%%%%%%%%%%%%%%%%%%%%%%%%%%%%%%%%%%%%%%%%%%%%%%%%%%%%%%%%%%%

\bibliography{main}

\end{document}